\documentclass[onecolumn]{ceurart}

\AtBeginDocument{%
  }

\copyrightyear{2025}
\copyrightclause{Copyright for this paper by its authors. Use permitted under Creative Commons License Attribution 4.0 International (CC BY 4.0).}
\conference{RS4SD – 1st Workshop on Recommender Systems for Sustainable Development using Responsible Nudging, Seoul, Korea}

\usepackage{multirow}

\usepackage[capitalise]{cleveref} 
\Crefname{section}{Sec.}{Secs.}
\Crefname{table}{Tab.}{Tabs.}

\usepackage{booktabs}
\usepackage{bbold}
\usepackage{amsmath}
\usepackage{algorithm, algorithmic}

\begin{document}

\title{Efficient Recommendations via Graph Coarsening and Label Propagation}

\author[12]{Alessandro Sbandi}[%
    email=alessandro.sbandi@telecomitalia.it,
    orcid=0009-0002-8804-2362
]

\author[1]{Federico Siciliano}[%
    email=siciliano@diag.uniroma1.it,
    orcid=0000-0003-1339-6983
]

\author[1]{Fabrizio Silvestri}[%
    email=fsilvestri@diag.uniroma1.it,
    orcid=0000-0001-7669-9055
]

\address[1]{Sapienza University of Rome, Rome, Italy}
\address[2]{TIM S.p.A.}

\begin{abstract}
Graph-based recommendations are widely adopted in real-world industrial applications. However, graphs in these systems often reach a massive scale, posing notable scalability and efficiency challenges.
This requires techniques that can effectively balance predictive quality with computational cost. One promising approach is graph coarsening, an adaptive graph reduction technique that offers a way to systematically construct smaller, yet structurally representative, versions of the original large-scale graphs.

In this work, we propose a flexible two-stage diffusion framework that combines graph coarsening with multi-step label propagation in the telecommunications domain. Domain-specific heuristics are applied to first aggregate nodes into meaningful communities, reducing graph size while preserving essential business-relevant relationships.
An initial diffusion process done by a Label Propagation Algorithm (LPA) or a Graph Neural Network (GNN) propagates labels across the coarsened graph to produce coarse-grained predictions. Finally, a second LPA within subgraphs generates the final recommendations for individual users.

On a real-world telecommunications dataset, when using LPA in both stages, our method achieves up to +24\% NDCG@5 over the full-graph LPA baseline. Incorporating a lightweight GNN in the first stage further boosts NDCG@5 by more than 50\%, but requires substantial training and inference time.
Through extensive experiments and a detailed ablation, we quantify these trade-offs and demonstrate that our coarsening-driven approach delivers an optimal balance between scalability, latency, and recommendation quality.
\end{abstract}

%

\maketitle


\section{Introduction}

Graph-structured data has become widespread across various domains, such as social networks \cite{fan2019graph, sharma2024survey}, biological networks \cite{wang2020leverage}, financial transactions \cite{wang2021review}, e-commerce platforms \cite{huang2004graph, xu2020product} because it naturally represents complex relationships between entities. In recommendation systems \cite{wu2022graph, gao2023survey}, user-item interactions form large bipartite graphs \cite{fan2019graph}, whose rich connectivity enables machine learning models to uncover latent patterns and deliver personalized recommendations \cite{pang2022heterogeneous}.
However, as industrial-scale datasets grow to millions or billions of nodes and edges \cite{dhulipala2023terahac}, this expressiveness comes at a steep computational cost: memory demands increase, training and inference times become prohibitively long, and graph algorithms face scalability bottlenecks \cite{sahu2017ubiquity,ju2024survey}.

Graph Neural Networks (GNNs) \cite{dong2021global, malewicz2010pregel} have  emerged as a powerful paradigm for recommendation, learning node embeddings through repeated message exchanges across edges. Yet even state-of-the-art large-scale GNNs rely on sampling techniques to handle these large-scale graphs, require heavy computational resources \cite{zhang2022understanding}, and struggle with problems like over-smoothing \cite{chen2020measuring} and capturing long-range dependencies \cite{li2018deeper}. These factors make handling truly massive graphs challenging in low-latency production settings and highlight the need for more scalable alternatives that maintain high-quality recommendations \cite{ju2024survey}.

Lightweight alternatives such as the Label Propagation Algorithm (LPA) \cite{xing2014node} bypass expensive gradient computations by diffusing information across the graph by iteratively updating node labels based on its neighbors. Compared to GNNs, LPA is computationally efficient and achieves competitive accuracy \cite{huang2020combining}. However, directly applying LPA to industrial-scale graphs still involves iterating over every edge multiple times, resulting in significant runtime and memory requirements. Naïve down-sampling of the graph can reduce costs, but it often severs critical links, destroying community structures or isolating hubs, which in turn degrades recommendation quality.

\begin{figure*}[t]
    \centering
    \includegraphics[width=\textwidth]{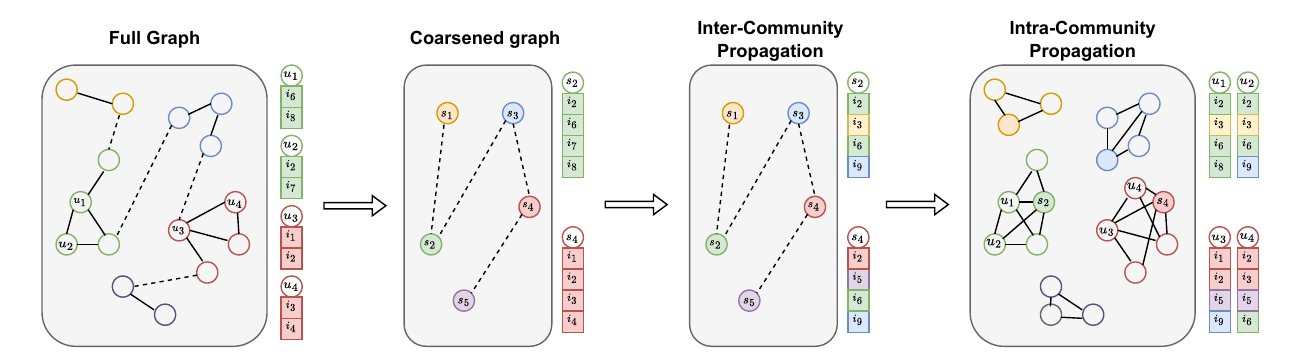}
    \caption{Overview of the proposed framework. On the left, the original full graph, where colors illustrate the business-driven rules used to group nodes into communities, while dashed lines highlight inter-community interactions. In the middle is shown the coarsened graph and inter-community propagation. On the right, the intra-community refinement stage is shown.}
    \label{img:framework_image}
\end{figure*}

Graph coarsening \cite{cai2021graph} offers a systematic approach to reduction, involving the aggregation of nodes to create a smaller, more computationally efficient representation of the original graph.
While coarsening can improve scalability \cite{huang2021scaling} and reduce noise \cite{10.1145/2872518.2889415}, it may lead to information loss, potentially impacting effectiveness. In fact, a propagation algorithm can diffuse labels efficiently on a coarsened graph, but recommendations become too coarse for fine-grained personalization.
In our industrial telecommunications setting, offers (e.g. additional services, bundled products) must be tailored to implicit user groups, that are not explicitly encoded in the raw data, and only made to users who do not already have these services.

To address these challenges, we propose a framework that integrates business-driven graph coarsening with at two-step propagation process. Our approach is designed for large-scale, real-world datasets and follow these steps: (i) first, users are grouped into super nodes that approximate real families by applying community detection techniques driven by business-specific domain rules, e.g shared billing and call frequency; 
(ii) either LPA or a lightweight GNN is applied on the coarsened graph to efficiently generate initial recommendations; (iii) a parallel label propagation step is performed within each subgraph to refine recommendations back down to the individual user level. This design combines the efficiency of reduction with the accuracy of localized propagation.

Offline experiments demonstrate the superiority of our proposed framework, achieving sub-second inference times and improving NDCG@5 by up to 24\% (LPA) and over 90\% (GNN) over a full-graph LPA baseline. Our contributions are:
\begin{itemize}
    \item We formalize a business-driven coarsening technique that uses business rules to uncover implicit family structures, preserving essential relationships and reducing the graph size by over 70\%.
    \item We have designed an efficient two-step propagation scheme that combines coarse-graph diffusion with fine-grained refinement to ensure low latency and high accuracy.
    \item We empirically validate our approach against full-graph LPA and alternative coarsening methods, demonstrating significant improvements in terms of scalability and recommendations quality.
\end{itemize}

\section{Methodology}
\cref{img:framework_image} presents our framework for scalable offer recommendations, which consists of three steps: 1) coarsening the interaction graph into communities, 2) propagating labels across communities, and 3) intra-community propagation.

Our data can be represented as an undirected user graph $G=(V, E, \textbf{Y})$, where $V = \{1,\dots,N\}$ is the set of nodes (users), $E = \{1,\dots,M\}$ is the set of edges (calls between users) and $\textbf{Y}\in \{0,1\}^{N\times L}$ is the label matrix, with $L$ denoting the number of unique labels (offers) and $\textbf{Y}_{ij}=1$ if user $i$ has accepted marketing offer $j$, $0$ otherwise.

\subsection{Graph Coarsening} A coarsened graph $G'=(V', E')$ contains $V'=\{1,\dots,N'\}$ super nodes and $E' = \{1,\dots,M'\}$ super edges, with $\textit{N'}<\textit{N}$. The coarsening process is represented by a surjective mapping $\textbf{C}:V\rightarrow V'$, where $\textbf{C} \in {\{0,1\}}^{N\times N'}$, ensuring each node $i$ in $V$ is mapped to exactly one super node in $V'$ (i.e. $\sum_{j=1}^{\textit{N'}} \textbf{C}_{ij} = 1$).


To ensure that the detected communities align with business goals, we incorporate domain-specific rules. The following heuristics indicate which users can be clustered together to capture implicit family and household structure:\\
\indent\textit{Interaction Frequency:} users who communicate more frequently based on call duration and number of calls per week.\\
\indent\textit{Surname:} users with the same surname.\\
\indent\textit{Caring:} users who frequently contacts customer support on behalf of someone else.\\
\indent\textit{Balance:} financial dependencies (e.g. topping up another user's mobile balance, charging someone else bills), especially for younger users\\
\indent\textit{Number of lines:} lines associated with a single user are redundant for personalized recommendation.

Users who satisfy at least one of these conditions and who comply with established marketing constraints (e.g. node degree and community size), are merged together to create a super node.

To create the new label matrix $\textbf{Y'}$, for each super node, we concatenate all label vectors $\textbf{y} \in \textbf{Y}$ of the nodes part of that community. Formally, given the mapping $\textbf{C}$,
the new label $\textbf{Y'}_j$ of community $j$ is computed  as $\textbf{Y'}_j=\mathbb{1}\left(\sum_{i=1}^N\textbf{C}_{ij}\textbf{Y}_j\geq1\right)$. This results in each community label $\textbf{Y'}_j$ containing all the offers activated by all of its users.
At this stage, super edges between communities are created based on inter-community interactions (e.g. calls between families).

\subsection{Inter-Community Propagation}
\label{sub:inter}
After graph coarsening, a first propagation is performed. We explore two algorithms: LPA and GNNs.

LPA offers an efficient method for information diffusion. It iteratively updates super nodes $v \in V^{'}$ labels based on predominant labels within their local neighborhoods.
The LPA is defined as:
\begin{equation}
\hat{\textbf{Y}}=\alpha \cdot \textbf{D}^{-1/2}\textbf{A}\textbf{D}^{-1/2}\textbf{Y}+(1-\alpha)\textbf{Y}
\nonumber
\end{equation}
where $\alpha$ is the propagation hyperparameter, $\textbf{A}$ is the adjacency matrix of the graph, and $\textbf{D}$ is the diagonal degree matrix with $\textbf{D}_{ii}$=$\sum_{j}\textbf{A}_{ij}$. 

When employing a GNN for the first propagation step, we adopt a more expressive graph representation to model interactions more accurately. Specifically, we transform the coarsened community-community graph $G'=(V',E')$ into a heterogeneous graph, denoted as $G_{h}$. This graph consists of two distinct node types: communities (the super nodes from $V'$) and items (the $L$ unique offers available in the catalog). The connectivity of $G_{h}$ is defined by two types of edges:

\begin{itemize}
    \item \textbf{Community-Community} edges, which correspond to the super-edges in $E'$.
    \item \textbf{Community-Item} edges, which capture the user-item interactions at the community level. 
\end{itemize}

To learn representations on this heterogeneous graph, we utilize the GraphSAGE architecture \cite{hamilton2017inductive}.
Let $\textbf{h}_{v}^{(0)}$ be the initial representations of super node $v\in V'$. In each layer $l$, an aggregated message $a_{v}^{(l)}$ is computed by applying an $\textit{AGG}$ function over the representations $\textbf{h}_{u}^{(l-1)}$ of $v$ neighboring super nodes $\textbf{u}\in N(v)$. $v$ representation is updated through $\textbf{h}_{v}^{(l)}=\sigma\left(\textbf{W}^{(l)}\cdot \textbf{a}_{v}^{(l)}+b^{(l)}\right)$, where $\textbf{W}^{(l)}$ and $b^{(l)}$ are learnable parameters. After $L$ layer, the final representation $\textbf{h}_{v}^{(L)}$ is obtained.

\subsection{Intra-Community Propagation} 
In this final step, we refine the initial recommendations obtained from the coarsened graph by leveraging the interactions within each community. Although the coarsened graph, or its heterogeneous counterpart, efficiently propagates information across large communities, it misses finer intra-community relationships. The recommendations generated from the inter-community step are uniform for all users within the same super-node, lacking the personalization required for individual level predictions. The intra-community step is designed to address this by refining these coarse-grained predictions and reintroducing user personalization.

This refinement is performed independently and in parallel for each community $c$ identified during the coarsening phase. We augment the original user-level subgraph of each community adding a community representative node $r_{c}$, containing  the aggregated label information from the previous propagation.
By connecting this representative node to all users in the community, we build a subgraph $G_{c,aug}$ that integrates both local (user-level) and global (community-level) information. Finally, we apply LPA over each subgraph to refine the recommendations, ensuring a more accurate and personalized output $\hat{\textbf{Y}}_{final}$. The presented framework is detailed in \cref{alg:framework}.

\begin{algorithm}[H]
\caption{Efficient Recommendations via Graph Coarsening and Label Propagation}
\label{alg:framework}
\begin{algorithmic}[1]
\STATE \textbf{Input:} Full user graph $G=(V, E, \textbf{Y})$; Coarsening business rules $\mathcal{R}$; Propagation algorithm $PA_{1st} \in \{\text{LPA, GNN}\}$
\STATE \textbf{Output:} Personalized recommendation scores $\hat{\textbf{Y}}_{final}$ for all users in $V$ \\
\textbf{Stage 1: Graph Coarsening}
    \STATE $E_{aux} \gets \emptyset$ \COMMENT{Initialize auxiliary edges for coarsening}
    \FORALL{user pair $(u, v) \in V \times V$}
        \IF{SatisfiesRules($(u, v), \mathcal{R}$) \textbf{and} MeetsConstraints($(u,v)$)}
            \STATE $E_{aux} \gets E_{aux} \cup \{(u, v)\}$
        \ENDIF
    \ENDFOR
    \STATE $\textbf{C} \gets \text{ConnectedComponents}(V, E_{aux})$ Generate community mapping matrix
    \STATE $G'=(V', E', \textbf{Y'}) \gets \text{CoarsenGraph}(G, \textbf{C})$ \COMMENT{Create coarsened graph}

    \vspace{0.5em}
    \textbf{Stage 2: Inter-Community Propagation}\\
    \IF{$PA_{inter}$ is GNN}
        \STATE $G_h \gets \text{CreateHeterogeneousGraph}(G')$
        \STATE $\mathbf{h}^{(L)}_{comm}, \mathbf{h}^{(L)}_{item} \gets \text{GraphSAGE}(G_h)$
        \STATE $\hat{\textbf{Y}}'_{scores} \gets \text{ComputeScores}(\mathbf{h}^{(L)}_{comm}, \mathbf{h}^{(L)}_{item})$ \COMMENT{Scores from GNN embeddings}
    \ELSE
        \STATE $\hat{\textbf{Y}}'_{scores} \gets \text{LPA}(G', \textbf{Y'})$ \COMMENT{Scores from LPA}
    \ENDIF

    \vspace{0.5em}
    \textbf{Stage 3: Intra-Community Propagation (in parallel)}\\
    \FORALL{community $c$ with super-node $v'_c \in V'$}
        \STATE $G_c \gets \text{ExtractSubgraph}(G, c)$ \COMMENT{Get original users and edges for community $c$}
        \STATE $r_c \gets \text{CreateRepresentativeNode}(\hat{\textbf{Y}}'_{scores}[c])$ \COMMENT{Node with global info}
        \STATE $G_{c,aug} \gets \text{AugmentSubgraph}(G_c, r_c)$ \COMMENT{Connect $r_c$ to all users in $G_c$}
        \STATE $\hat{\textbf{Y}}_{final} \gets \text{LPA}(G_{c,aug})$ \COMMENT{Refine recommendations locally}
    \ENDFOR
    \RETURN $\hat{\textbf{Y}}_{final}$
\end{algorithmic}
\end{algorithm}

\section{Experiments}
We conduct comprehensive experiments to demonstrate the effectiveness of the proposed method. Specifically, we aim to answer the following research questions:
\begin{itemize}
    \item \textbf{Effectiveness (RQ1):} How does our graph coarsening, compare in terms of recommendation quality, to (i) full-graph label propagation and (ii) alternative coarsening methods?
    \item \textbf{Structural integrity (RQ2):} To what extent do different coarsening strategies preserve the topological properties (e.g. community structure, clustering and degree distribution) of the original graph?
    \item \textbf{Efficiency and Scalability (RQ3):} How does the proposed approach perform in terms of computational efficiency and scalability when compared to full-graph propagation and other coarsening approaches?
     \item \textbf{Ablation (RQ4):} Which LPA design choices (number of propagation layers $L$ and propagation hyperparameter $\alpha$) drive performance and efficiency gains?
\end{itemize}
\subsection{Experimental Setup}
\subsubsection{Dataset}
We employ a real-world industrial dataset, containing more than 13 million users, collected from marketing campaigns run by TIM, an Italian telecommunications company, containing more than 13 million users. Specifically, it captures instances where customer care agents, during phone calls, suggest multiple products or services to a customer in a single engagement. This characteristic is central in the evaluation process, as discussed in \cref{subsub:eval}, because agents usually present a list of recommendations.

The dataset spans from January to September 2024 and captures user interactions with promotional offers. To ensure a realistic evaluation, we split the data chronologically: the last month is used as test set, and the previous months as training set. \cref{tab:dataset_summary} provides comprehensive details about train and test split statistics.

\begin{table}[ht]
\centering
\caption{Summary of Dataset Statistics}
\label{tab:dataset_summary}
\begin{tabular}{l|r|r}
\toprule
& Train & Test \\
\midrule
Unique users buying & 2,552,477 & 736,648\\ 
Total items bought & 4,818,165 & 1,070,679\\ 
\midrule
\% of users with $\geq$ 1 items bought & $18.49$\% & $5.34$\%\\ 
\% of users with $\geq$ 2 items bought & $7.34\%$ & $1.42$\%\\ 
\% of users with $\geq$ 3 items bought & $3.62\%$ & $0.53$\%\\ 
\bottomrule
\end{tabular}
\end{table}

The item catalog within our industrial recommendation dataset covers a diverse range of offers:
\begin{itemize}
    \item \textbf{Telecommunications Services:} including voice and Wi-Fi plans.
    \item \textbf{TV \& Streaming Subscriptions:} such as DAZN, Netflix, Amazon Prime, Disney+, and TIMVISION.
    \item \textbf{Promotional Discounts:} such as percentage-off deals or limited-time offer.
    \item \textbf{Smart Home Devices:} for example, Google Nest or security cameras.
    \item \textbf{Network Equipment:} including routers and related hardware.
\end{itemize}

The nature of offers allowing for collective access or use by a group (such as a family) has a dual implication for our recommendation task. Firstly, it can lead to lower individual adoption rates and contribute to dataset sparsity. Secondly, this makes identifying family members (or communities) particularly valuable for tailoring recommendations, as it's often redundant or ineffective to recommend the same item to multiple people in the same family.

Each of the $561$ unique items represents a distinct promotional offer or service package. We focus on purchases as the primary signal of user intent. This naturally leads to a sparse dataset, with an overall interaction density of $0.06\%$ in the training set. This sparsity highlights one of the challenges of industrial recommendation.

\subsubsection{Baseline Models}
We compare the proposed framework with three coarsening baselines:
\begin{itemize}
    \item \textit{Users}: the full user's graph, where each node represents an individual user and edges denote direct interactions. 
    \item \textit{Unique}: a coarsened version where super nodes are created based on unique label combinations.
    \item \textit{Louvain}: a coarsened version where the Louvain algorithm \cite{blondel2008fast} is used to partition the original graph into communities.
\end{itemize}

To ensure consistency in the evaluation, for all models, we follow the same methodology as in the proposed framework: PA is first applied to the coarsened graph, then LPA within each community.

\subsubsection{Implementation Details}
\label{subsub:implem}
As mentioned in \cref{sub:inter}, we implemented a GraphSAGE model for the GNN-based inter-community propagation. The model was trained to perform link prediction on the heterogeneous graph $G_{h}$, with the objective of predicting \textbf{Community-Item} edges. Specifically, the model was trained using a binary cross-entropy loss function, and key hyperparameters (learning rate, hidden dimension, number of layers, negative samples, batch size) were tuned using a Bayesian optimization on the validation set \cite{akiba2019optuna}.

\subsubsection{Evaluation Metrics}
\label{subsub:eval}
To evaluate the effectiveness of our model, we investigate three areas.
First, we measure the quality of the recommendation using standard ranking metrics: Precision@K, NDCG@K, Recall@K and Mean Average Precision (MAP).
To evaluate the structural preservation of the coarsened graphs, we measure Degree Similarity (DS), Average Clustering Coefficient (AC), Connected Components Similarity (CCS), and Edge Density (ED), along with a comparison of the node count. This helps us determine if the coarsening process retains the key features of the original graph. The DS and CCS are computed using the formula $1/(1+d(x, y))$, where $d$ is the Wasserstein distance, and $x$ and $y$ are the lists of node degrees or connected component sizes for the original and coarsened graphs, respectively.
Finally, we compare runtime efficiency by recording LPA execution time, to show how scalable and fast our approach is when handling large graphs.

Given that customer care agents typically propose at least five items during these interactions, evaluating the effectiveness of their call recommendations requires to consider and focus on K=5. This is precisely why NDCG@5 is our primarily evaluation metric. It allows us to measure how well the most relevant items are prioritized within the top 5 positions, which is crucial in a scenario where an agent offers a limited set of options during a brief interaction.
\subsection{Effectiveness (RQ1)}

\looseness -1 

\cref{tab:results} reports the results for the complete set of metrics of our proposed framework against several baselines. The comparison includes our coarsening business-driven strategy and alternative methods, each evaluated with LPA or GNN as the first propagation algorithm, and the full-graph LPA baseline. A critical initial observation is the Out Of Memory (OOM) encountered when attempting to apply the GNN to the original full graph, underlining the necessity of graph coarsening. All of these use the best configuration found (see \cref{sec:ablation} for LPA). Since customer agents typically recommend up to five offers from the recommendation list, we set the primary evaluation goal to $\textit{K}=5$.

\begin{table*}[h]
    \centering
    \resizebox{\textwidth}{!}{
    \begin{tabular}{cc | l lll lll lll lll}
        \toprule
        \multirow{3}{*}{1st PA} & \multirow{3}{*}{Model} & \multirow{3}{*}{MAP} 
        & \multicolumn{3}{c}{@3} & \multicolumn{3}{c}{@5} 
        & \multicolumn{3}{c}{@10} & \multicolumn{3}{c}{@20} \\
        \cmidrule(lr){4-6} \cmidrule(lr){7-9} \cmidrule(lr){10-12} \cmidrule(lr){13-15}
        & & & \multicolumn{1}{c}{P} & \multicolumn{1}{c}{NDCG} & \multicolumn{1}{c}{R} & \multicolumn{1}{c}{P} & \multicolumn{1}{c}{NDCG} & \multicolumn{1}{c}{R} & \multicolumn{1}{c}{P} & \multicolumn{1}{c}{NDCG} & \multicolumn{1}{c}{R} & \multicolumn{1}{c}{P} & \multicolumn{1}{c}{NDCG} & \multicolumn{1}{c}{R} \\
        \midrule
        \multirow{4}{*}{LPA} & \textit{Users} & 0.0220& \underline{0.0155} & 0.0260& \underline{0.0275} & \underline{0.0112} & \underline{0.0275} & \underline{0.0318} & \underline{0.0062} & \underline{0.0282} & \underline{0.0339} & \underline{0.0031} & \underline{0.0282} & 0.0341 \\
        & \textit{Unique} & 0.0181 & 0.0105 & 0.0191 & 0.0207 & 0.0080 & 0.0211 & 0.0260 & 0.0051& 0.0236 & 0.0330& 0.0028& 0.0243& \underline{0.0357}  \\
        & \textit{Louvain} & \underline{0.0234} & 0.0134& \underline{0.0270} & 0.0268 & 0.0084& 0.0270 & 0.0281& 0.0043& 0.0271& 0.0285& 0.0021& 0.0271& 0.0286 \\
        & \textit{Ours} & \textbf{0.0281} & \textbf{0.0182} & \textbf{0.0313}& \textbf{0.0332} & \textbf{0.0138}& \textbf{0.0341}& \textbf{0.0405} & \textbf{0.0084}& \textbf{0.0367}& \textbf{0.0481}& \textbf{0.0043}& \textbf{0.0372}& \textbf{0.0498}\\
        \midrule
        \multirow{4}{*}{GNN} & \textit{Users} & OOM & OOM & OOM & OOM & OOM & OOM & OOM & OOM & OOM & OOM & OOM & OOM & OOM\\
        & \textit{Unique} & 0.0305 & 0.0172 & 0.0318 & 0.0321 & \underline{0.0124} & 0.0352 & 0.0393 & \underline{0.0078} & 0.0382 & 0.0455 & 0.0043 & 0.0394 & \underline{0.0491} \\
        & \textit{Louvain} & \underline{0.0390} & \underline{0.0181} & \underline{0.0373} & \textbf{0.0489} & 0.0121 & \underline{0.0392} & \textbf{0.0532} & 0.0095 & \underline{0.0498} & \textbf{0.0859} & \textbf{0.0058} & \textbf{0.0543} & \textbf{0.1026}\\
        & \textit{Ours} & \textbf{0.0434} & \textbf{0.0312} & \textbf{0.0552} & \underline{0.0461} & \textbf{0.0193} & \textbf{0.0528} & \underline{0.0471} & \textbf{0.0098} & \textbf{0.0524} & \underline{0.0478} & \underline{0.0049} & \underline{0.0524} & 0.0478  \\
        \bottomrule
    \end{tabular}}
    \caption{Performance metrics averaged over $3$ runs for baselines. Standard deviation is not reported due to space constraints, but is always less than $0.001$. PA = Propagation Algorithm, OOM = Out Of Memory. \textbf{Bold} indicates the best metric with those parameters, \underline{underlined} are second best.}
    \label{tab:results}
\end{table*}

When LPA is applied as the first PA, our approach achieves the highest recommendation quality consistently across all metrics. Notably, our method improves NDCG@5 by $24\%$ over the full-graph propagation baseline.

Using a GNN as the first PA notably enhances performance, particularly when combined with our coarsening strategy. Our approach achieves overall best results across several metrics. It improves the $NDCG@5$ by $54\%$ over its LPA counterpart and by $35\%$ over the GNN trained on $\textit{Louvain}$ coarsened graph. The latter generally stands as the second best method, achieving always highest recall and best metrics @20.

In summary, these results confirm that graph coarsening is indispensable for applying GNNs to large industrial-scale graphs. Moreover, $\textit{Ours}$ business-driven coarsening approach is consistently having high performance with both (LPA and GNN) PA. GNN in the first propagation step achieves the highest recommendation quality, significantly outperforming LPA approaches.

\subsection{Structural Integrity (RQ2)}

\cref{tab:graph_properties} reports the five structural metrics for each coarsening strategy compared to the original user–item graph.
Together, these metrics help quantify how faithfully each method preserves global and local structural properties.

Our approach yields an AC that most closely matches that of the original graph. The low clustering coefficient similarity reflects the sparse and loosely connected nature of user-item interactions within the generated communities, which is consistent with typical user behavior on large-scale platforms. This low clustering coefficient aligns with our goal of preserving realistic connectivity patterns without introducing artificial density. Similarly, the edge density (ED) of our coarsened graph remains low, indicating that the compression process preserves the original structure's natural sparsity, avoiding excessive over-connection among super nodes. At the same time, our method achieves moderate compression (in terms of node count), avoiding the extreme reductions that may compromise structural fidelity.

\begin{table}[h!]
    \centering
    \begin{tabular}{lcccccc}
        \toprule
        \textbf{Graph} & \textbf{DS} & \textbf{AC} & \textbf{CCS} & \textbf{ED} & \textbf{NN} \\
        \midrule
        \textit{Users}        & - & $6.287\cdot10^{-7}$ & - & $2.839\cdot10^{-7}$ & 13.80 \\
        \midrule
        \textit{Unique}  & 0.263 & 0.327 & $2.820\cdot10^{-4}$ & $3.312\cdot10^{-5}$ & 0.14 \\
        \textit{Louvain} & \textbf{0.587} & \underline{0.984} & \textbf{5.152$\cdot$10$^{-2}$} & \underline{$1.976\cdot10^{-6}$} & 0.49 \\
        \textit{Ours}      & \underline{0.485} & \textbf{1.070$\cdot$10$^{-6}$} & \underline{$1.647\cdot10^{-2}$} & \textbf{9.105$\cdot$10$^{-7}$} & 4.27 \\
        \bottomrule
    \end{tabular}
    \caption{Comparison of graph properties across different methods: degree similarity (DS), average clustering (AC), connected components similarity (CCS), edge density (ED), and number of nodes (NN) in millions.}
    \label{tab:graph_properties}
\end{table}

\newpage

The \textit{Louvain} method produces a more compact graph than ours, and it exhibits the highest DS and CCS scores, indicating that it doesn't distort the original component structure and degree distribution heavily. Despite a low ED value — suggesting nodes with poor connectivity — the AC deviates greatly from the original graph’s, limiting \textit{Louvain} effectiveness.

On the other hand, the \textit{Unique} coarsening strategy achieves the most aggressive reduction in node count, resulting in a very compact graph. It yields the lowest DS and CCS, indicating that it doesn't preserve some macro-structural properties of the original graph. However, this comes at the cost of excessive edge density and lowest clustering coefficient similarity, both of which suggest overly tight communities that may inhibit meaningful diffusion and lead to unstable performance (as observed in RQ1).

Overall, our coarsening framework strikes the best balance between graph compression and structural fidelity.

\begin{figure*}[!ht]
    \centering
    \includegraphics[width=0.95\textwidth]{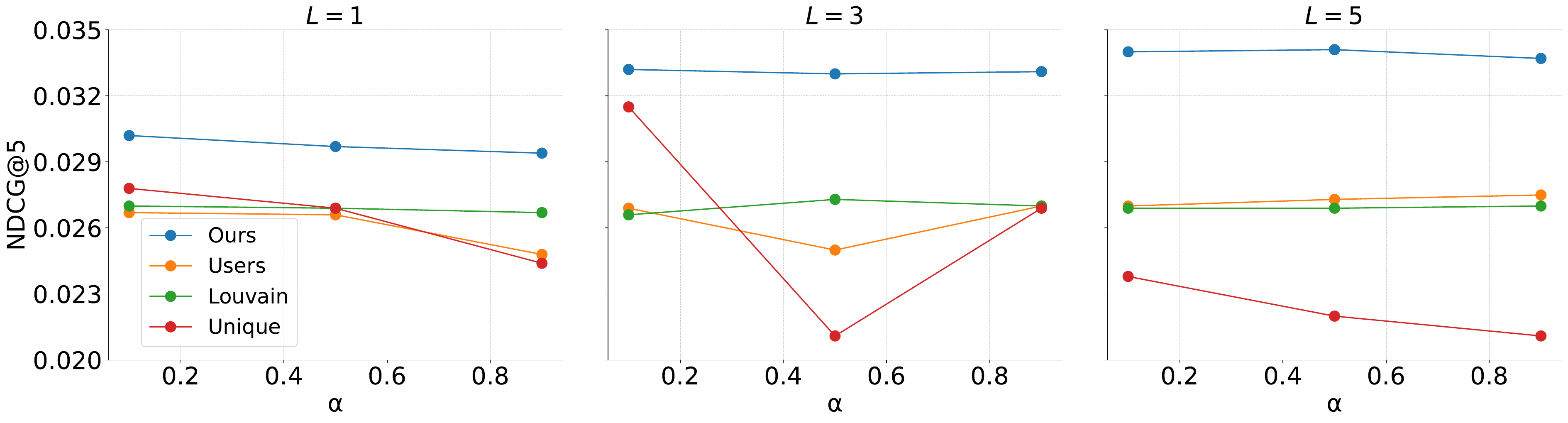}
    \caption{NDCG@5 values for all coarsening methods, varying the number of layers $L$, with corresponding $\alpha$ values indicated.}
    \label{fig:ndcg@5}
\end{figure*}

\subsection{Efficiency and Scalability (RQ3)}

\cref{tab:graph_complexity} summarizes, for each method, the graph size (nodes and edges) at each propagation step and the corresponding runtimes for both coarsening and propagation steps. The full-graph baseline ($\textit{Users}$), which involves performing a single propagation on an original graph comprising roughly 14 million nodes and 10 million edges, taking significantly longer ($>$17 minutes) than all the other methods. Crucially, attempting to run GNN propagation on this full graph results in Out Of Memory (OOM), highlighting the necessity of coarsening for GNN-based approaches. 

By contrast, our business-driven coarsening achieves a minimal coarsening time ($34$s), thanks to lightweight heuristics that can be precomputed offline. The resulting coarsened graph contains 4 million nodes and 6 million edges, and undergoes two LPA propagation steps in nine minutes: the first on the reduced graph and the second in parallel across smaller user-level subgraphs. On the other side, GNN propagation is the slowest with approximately 13 minutes.

The \textit{Unique} strategy completes the coarsening process in around 4 minutes), producing the smaller graph (140 thousand nodes and 270 thousand edges). While its first propagations are the fastest ($\sim$5 minutes LPA, $\sim$10 minutes GNN), this advantage is reversed in the second step, where it takes over $4$ seconds per graph, the longest of all methods, due denser communities.
While Louvain coarsening produces well-balanced partitions, it is the slowest to run, taking over two hours. Though its propagation steps are marginally faster than ours, the upfront cost dominates its total runtime and, ultimately, our approach remains competitive in terms of total LPA and GNN runtime while achieving superior recommendation performance, as shown in \cref{tab:results}.

From a theoretical point of view, from \cite{raghavan2007near,wang2020label,traag2023large}, LPA algorithm has a running time of $\mathcal{O}(TM)$, where $T$ is the number of iterations and $M$ is the number of edges in the graph.
The number of iterations can either be fixed or left to convergence. However, no theoretical proof of LPA convergence time has been found \cite{traag2023large} except for special cases. Empirical studies suggest that a few iterations are usually sufficient to stabilize the labels \cite{raghavan2007near}. Reported complexity also depends on LPA variant, with studies citing $\mathcal{O}(N log^2 N)$ \cite{clauset2004finding} and $\mathcal{O}(M+N)$ \cite{wu2004finding}.
GNNs typically exhibit higher computational complexity: with $L$ layers, a graph with $N$ nodes, and $M$ edges, the complexity is $\mathcal{O}(L\cdot (M + N))$ \cite{wu2020comprehensive}. This factor inherently make GNNs more resource-intensive compared to LPA. 

In our case, the first step, whether LPA or GNN, is expected to be the most computationally intensive due to the large scale of the coarsened graph, which contains millions of nodes and edges. The GNN runtimes clearly demonstrate a higher computational demand than LPA on the same coarsened graph. Conversely, the second LPA step operates on much smaller community graphs, containing only tens or hundreds of nodes, making it computationally light. An additional advantage of our approach is that the second LPA step can be efficiently parallelized. As it operates at the user level, it can be performed online at inference time, while the first propagation step can be precomputed offline. This design ensures that our method remains scalable and operates within sub-second inference times.

\begin{table*}[!ht]
    \centering
    \resizebox{\textwidth}{!}{
    \begin{tabular}{ll|cccc|ccc}
        \toprule
        \multicolumn{2}{c|}{\textbf{Coarsening}} & \multicolumn{4}{c|}{\textbf{First Propagation}} & \multicolumn{3}{c}{\textbf{Second LPA (per community)}}\\
        \textbf{Approach} & \textbf{Runtime} & \textbf{Nodes (M)} & \textbf{Edges (M)} & \textbf{LPA Runtime} & \textbf{GNN Runtime} & \textbf{Nodes} & \textbf{Edges} & \textbf{LPA Runtime} \\
        \midrule
        \textit{Users} & \textbf{0s} & 13.80 & 9.65 & 17m41s$\pm$30s & $\textit{OOM}$ &- & - & - \\
        \midrule
        \textit{Unique}  & 3min53s$\pm$40s & 0.14 & 0.27 & \textbf{5m07s$\pm$50s} & \textbf{10m39s$\pm$24s} & 99.57 & 463.57 & 4457ms$\pm$264ms \\
        \textit{Louvain} & 2h12min$\pm$2m & 0.49 & 0.26 & 6m47s$\pm$30s & 11m02s$\pm$35s & 29.16 & 32.65 & 1571ms$\pm$423ms\\
        \textit{Ours}      & \textbf{34s$\pm$5s} & 4.27 & 6.36 & 9m01s$\pm$47s & $12m59s\pm$14s & 4.23 & 7.68 & \textbf{358ms$\pm$24ms} \\
        \bottomrule
    \end{tabular}
    }
    \caption{Comparison of graph statistics across the three steps, including number of nodes, edges and runtime. The best runtime for each step is highlighted in \textbf{bold}.}
    \label{tab:graph_complexity}
\end{table*}

\subsection{Ablation (RQ4)}
\label{sec:ablation}

To analyze the effect of the main hyperparameters on recommendation accuracy, we perform an ablation study by varying the number of propagation layers $\textit{L} \in \{1,3,5\}$, and damping parameter $\alpha \in \{0.1,0.5,0.9\}$. Due to the computational complexity associated with GNN training, we conduct this ablation study specifically using LPA as our propagation algorithm. For each configuration, we evaluate all coarsening strategies against the full graph baseline. 

As shown in \cref{tab:ablation}, our coarsening approach consistently outperforms all baselines across different hyperparameter settings. In particular, it shows a clear upward trend in performance as the number of propagation layers $L$ increases, while the performance of the other models remain largely stable. This is also observable in \cref{fig:ndcg@5}. Moving from a single propagation step to five, yields and average NDCG@5 gain of over $17$\%, regardless of the chosen $\alpha$.

In contrast, the alternative coarsening methods and full-graph propagation remain largely insensitive to changes in $L$ or $\alpha$. As can be seen in \cref{fig:ndcg@5}, their NDCG@5 curves are essentially flat, indicating that deeper or more aggressive propagation does not result in significant improvements to either the original or differently coarsened graphs. The only exception is \textit{Unique}, which shows a high sensitivity to hyperparameter variation - its performance deteriorates at $L=5$ and fluctuates unpredictably with $\alpha$ when $L=3$.

Based on these results, we set the optimal hyperparameters to $\textit{L}=5$ and $\alpha=0.5$, as this configuration gives the highest NDCG@5.

\begin{table*}[h]
    \centering
    \resizebox{\textwidth}{!}{
    \begin{tabular}{c c c | l lll lll lll lll}
        \toprule
        \multirow{3}{*}{\textit{L}} & \multirow{3}{*}{$\alpha$} & \multirow{3}{*}{Model} & \multirow{3}{*}{MAP} 
        & \multicolumn{3}{c}{@3} & \multicolumn{3}{c}{@5} 
        & \multicolumn{3}{c}{@10} & \multicolumn{3}{c}{@20} \\
        \cmidrule(lr){5-7} \cmidrule(lr){8-10} \cmidrule(lr){11-13} \cmidrule(lr){14-16}
        
        & & & & P & NDCG & R & P & NDCG & R & P & NDCG & R & P & NDCG & R \\
        
        \midrule
        
        \multirow{12}{*}{1}  
                            & \multirow{4}{*}{0.1} & \textit{Users} & 0.0213 & \underline{0.0152} & 0.0256 & 0.0267 & \underline{0.0109} & 0.0267 & 0.0304 & 0.0059 & \underline{0.0272} & 0.0319 & 0.0030 & 0.0272 & 0.0320  \\
                            & & \textit{Unique} & \underline{0.0240} & 0.0134 & 0.0243 & 0.0266 & \underline{0.0109} & \underline{0.0278} & \textbf{0.0354} & \textbf{0.0069} & \textbf{0.0309} & \textbf{0.0442} & \textbf{0.0036} & \textbf{0.0315} & \textbf{0.0462}  \\
                            & & \textit{Louvain} & 0.0235 & 0.0132 & \underline{0.0270} & \underline{0.0269} & 0.0084 & 0.0270 & 0.0282 & 0.0043 & 0.0270 & 0.0287 & 0.0021 & 0.0271 & 0.0287  \\
                            & & \textit{Ours} & \textbf{0.0243} &  \textbf{0.0168} & \textbf{0.0287} & \textbf{0.0302} & \textbf{0.0121} & \textbf{0.0302} & \underline{0.0347} & \underline{0.0066} & \textbf{0.0309} & \underline{0.0368} & \underline{0.0033} & \underline{0.0309} & \underline{0.0369} \\
        \cmidrule(lr{0.5pt}){2-16}
                            & \multirow{4}{*}{0.5} & \textit{Users} & 0.0212 & 0.0152 & 0.0255 & 0.0266 & 0.0109 & 0.0266 & 0.0302 & \underline{0.0060} & 0.0271 & \underline{0.0317} & \underline{0.0029} & 0.0271 & 0.0319 \\
                            & & \textit{Unique} & 0.0164 & \underline{0.0158} & 0.0245 & 0.0263 & \underline{0.0116} & \underline{0.0269} & \underline{0.0327} & 0.0054 & \underline{0.0295} & 0.0297 & \underline{0.0029} & \underline{0.0301} & \underline{0.0421}  \\
                            & &\textit{ Louvain} & 0.0234 & 0.0132 & \underline{0.0269} & \underline{0.0268} & 0.0084 & \underline{0.0269} & 0.0282 & 0.0043 & 0.0270 & 0.0286 & 0.0021 & 0.0270 & 0.0286  \\
                            & & \textit{Ours} & \textbf{0.0239} & \textbf{0.0166} & \textbf{0.0282} & \textbf{0.0297} & \textbf{0.0120} & \textbf{0.0297} & \textbf{0.0342} & \textbf{0.0065} & \textbf{0.0304} & \textbf{0.0363} & \textbf{0.0033} & \textbf{0.0304} & \textbf{0.0364} \\
        \cmidrule(lr{0.5pt}){2-16}
                            & \multirow{4}{*}{0.9} & \textit{Users} & 0.0215 & 0.0122 & 0.0248 & 0.0244 & 0.0077 & 0.0248 & 0.0256 & 0.0039 & 0.0248 & 0.0261 & 0.0019 & 0.0248 & 0.0261 \\
                            & & \textit{Unique} & 0.0138 & 0.0120 & 0.0218 & 0.0239 & \underline{0.0095} & 0.0244 & \underline{0.0303} & \underline{0.0063} & \underline{0.0272} & \underline{0.0333} & \underline{0.0027} & 0.0278 & 0.0405  \\
                            & & \textit{Louvain} & \underline{0.0233} & \underline{0.0132} & \underline{0.0269} & \underline{0.0264} & 0.0082 & \underline{0.0267} & 0.0274 & 0.0041 & 0.0267 & \underline{0.0278} & 0.0021 & 0.0268 & 0.0279  \\
                            & & \textit{Ours} & \textbf{0.0236} &  \textbf{0.0165} & \textbf{0.0282} & \textbf{0.0294} & \textbf{0.0118} & \textbf{0.0294} & \textbf{0.0333} & \textbf{0.0064} & \textbf{0.0299} & \textbf{0.0349} & \textbf{0.0032} & \textbf{0.0299} & \underline{0.0350} \\
        \midrule
        
        \multirow{12}{*}{3}  
                            & \multirow{4}{*}{0.1} & \textit{Users} & 0.0215 & \underline{0.0152} & 0.0257 & 0.0269 & 0.0110 & 0.0269 & 0.0307 & 0.0060 & 0.0275 & 0.0326 & \underline{0.0030} & 0.0275 & 0.0327 \\
                            & & \textit{Unique} & \underline{0.0266} & 0.0149 & \underline{0.0274} & \underline{0.0298} & \underline{0.0122} & \underline{0.0315} & \textbf{0.0399} & \textbf{0.0079} & \textbf{0.0353} & \textbf{0.0505} & \textbf{0.0040} & \textbf{0.0358} & \textbf{0.0521}  \\
                            & & \textit{Louvain} & 0.0232 & 0.0131 & 0.0266 & 0.0264 & 0.0082 & 0.0266 & 0.0277 & 0.0042 & 0.0266 & 0.0282 & 0.0021 & 0.0266 & 0.0282  \\
                            & & \textit{Ours} & \textbf{0.0271} &  \textbf{0.0179} & \textbf{0.0307} & \textbf{0.0326} & \textbf{0.0135} & \textbf{0.0332} & \underline{0.0393} & \underline{0.0078} & \underline{0.0351} & \underline{0.0448} & \textbf{0.0040} & \underline{0.0354} & \underline{0.0458} \\
        \cmidrule(lr{0.5pt}){2-16}
                            & \multirow{4}{*}{0.5} & \textit{Users} & 0.0215 & 0.0123 & 0.0248 & 0.0247 & 0.0078 & 0.0250 & 0.0263 & 0.0040 & 0.0251 & 0.0268 & 0.0020 & 0.0251 & 0.0269 \\
                            & & \textit{Unique} & 0.0175 & 0.0105 & 0.0184 & 0.0205 & \underline{0.0085} & 0.0211 & 0.0273 & \underline{0.0054} & 0.0237 & \underline{0.0345} & \underline{0.0028} & 0.0242 & \underline{0.0363}  \\
                            & & \textit{Louvain} & \underline{0.0238} & \underline{0.0134} & \underline{0.0273} & \underline{0.0269} & 0.0084 & \underline{0.0273} & \underline{0.0283} & 0.0043 & \underline{0.0273} & 0.0289 & 0.0022 & \underline{0.0273} & 0.0289  \\
                            & & \textit{Ours} & \textbf{0.0270} &  \textbf{0.0179} & \textbf{0.0308} & \textbf{0.0325} & \textbf{0.0132} & \textbf{0.0330} & \textbf{0.0386} & \textbf{0.0077} & \textbf{0.0348} & \textbf{0.0437} & \textbf{0.0039} & \textbf{0.0350} & \textbf{0.0445} \\
        \cmidrule(lr{0.5pt}){2-16}
                            & \multirow{4}{*}{0.9} & \textit{Users} & 0.0216 & \underline{0.0153} & 0.0257 & \underline{0.0270} & \underline{0.0110} & \underline{0.0270} & 0.0309 & 0.0060 & 0.0276 & 0.0327 & 0.0030 & 0.0276 & 0.0329 \\
                            & & \textit{Unique} & 0.0231 & 0.0125 & 0.0238 & 0.0262 & 0.0098 & 0.0269 & \underline{0.0339} & \underline{0.0062} & \underline{0.0298} & \underline{0.0421} & \underline{0.0032} & \underline{0.0303} & \underline{0.0438}  \\
                            & & \textit{Louvain} & \underline{0.0234} & 0.0135 & \underline{0.0271} & 0.0268 & 0.0085 & \underline{0.0270} & 0.0281 & 0.0043 & 0.0271 & 0.0285 & 0.0022 & 0.0271 & 0.0286  \\
                            & & \textit{Ours} & \underline{0.0182} & \textbf{0.0179} & \textbf{0.0308} & \textbf{0.0327} & \textbf{0.0134} & \textbf{0.0331} & \textbf{0.0391} & \textbf{0.0078} & \textbf{0.0350} & \textbf{0.0446} & \textbf{0.0040} & \textbf{0.0352} & \textbf{0.0454} \\
       \midrule
       
       \multirow{12}{*}{5}  
                           & \multirow{4}{*}{0.1} & \textit{Users} & 0.0215 & \underline{0.0153} & 0.0257 & \underline{0.0270} & \underline{0.0110} & 0.0269 & \underline{0.0309} & \underline{0.0060} & \underline{0.0276} & 0.0328 & 0.0030 & \underline{0.0276} & 0.0329 \\
                           & & \textit{Unique} & 0.0199 & 0.0120 & 0.0211 & 0.0234 & 0.0095 & 0.0240 & 0.0308 & 0.0059 & 0.0264 & \underline{0.0378} & \underline{0.0031} & 0.0272 & \underline{0.0403}  \\
                           & & \textit{Louvain} & \underline{0.0235} & 0.0133 & \underline{0.0270} & 0.0269 & 0.0084 & \underline{0.0270} & 0.0283 & 0.0043 & 0.0270 & 0.0287 & 0.0021 & 0.0271 & 0.0288  \\
                           & & \textit{Ours} & \textbf{0.0282} & \textbf{0.0182} & \textbf{0.0314} & \textbf{0.0334} & \textbf{0.0138} & \textbf{0.0341} & \textbf{0.0405} & \textbf{0.0084} & \textbf{0.0368} & \textbf{0.0481} & \textbf{0.0043} & \textbf{0.0373} & \textbf{0.0499} \\
       \cmidrule(lr{0.5pt}){2-16}
                           & \multirow{4}{*}{0.5} & \textit{Users} & 0.0219 & \underline{0.0155} & 0.0260 & \underline{0.0274} & \underline{0.0112} & \underline{0.0273} & \underline{0.0315} & \underline{0.0062} & \underline{0.0280} & \underline{0.0335} & \underline{0.0031} & \underline{0.0281} & 0.0336 \\
                           & & \textit{Unique} & 0.0194 & 0.0111 & 0.0205 & 0.0221 & 0.0086 & 0.0227 & 0.0280 & 0.0054 & 0.0252 & 0.0349 & 0.0029 & 0.0260 & \underline{0.0378}  \\
                           & & \textit{Louvain} & \underline{0.0234} & 0.0132 & \underline{0.0269} & 0.0264 & 0.0083 & 0.0268 & 0.0276 & 0.0042 & 0.0268 & 0.0280 & 0.0021 & 0.0269 & 0.0281  \\
                           & & \textit{Ours} & \textbf{0.0283} & \textbf{0.0182} & \textbf{0.0314} & \textbf{0.0333} & \textbf{0.0138} & \textbf{0.0342} & \textbf{0.0406} & \textbf{0.0084} & \textbf{0.0368} & \textbf{0.0482} & \textbf{0.0043} & \textbf{0.0373} & \textbf{0.0499} \\
       \cmidrule(lr{0.5pt}){2-16}
                           & \multirow{4}{*}{0.9} & \textit{Users} & 0.0218 & \underline{0.0154} & 0.0259 & \underline{0.0273} & \underline{0.0111} & \underline{0.0273} & \underline{0.0315} & \underline{0.0062} & \underline{0.0279} & \underline{0.0335} & \underline{0.0031} & \underline{0.0280} & 0.0337 \\
                           & & \textit{Unique} & 0.0175 & 0.0101 & 0.0184 & 0.0202 & 0.0077 & 0.0204 & 0.0253 & 0.0050 & 0.0231 & 0.0330 & 0.0027 & 0.0238 & \underline{0.0357}  \\
                           & & \textit{Louvain} & \underline{0.0233} & 0.0132 & \underline{0.0268} & 0.0267 & 0.0083 & 0.0268 & 0.0280 & 0.0042 & 0.0269 & 0.0284 & 0.0021 & 0.0269 & 0.0285  \\
                           & & \textit{Ours} & \textbf{0.0282} & \textbf{0.0181} & \textbf{0.0312} & \textbf{0.0332} & \textbf{0.0138} & \textbf{0.0340} & \textbf{0.0406} & \textbf{0.0085} & \textbf{0.0368} & \textbf{0.0487} & \textbf{0.0044} & \textbf{0.0374} & \textbf{0.0505} \\
        \bottomrule
    \end{tabular}}
    
    \caption{Performance metrics with baselines for different layers and alpha values. $\textbf{Bold}$ and $\underline{underlined}$ indicate the best and second-best metric with those parameters, respectively.}
    \label{tab:ablation}
\end{table*}

\section{Conclusion}
In this paper, we have introduced a novel graph coarsening framework for industrial-scale recommendation systems that incorporates business-driven rules to improve scalability. The framework involves a two-stage propagation: inter-community propagation (LPA or GNN-based) is applied to the coarsened graph to efficiently spread information across communities, and intra-community propagation refines the initial recommendations within each community.
This method effectively utilizes user interactions to enhance the recommendation of family offers and provides more scalable recommendation updates, which is crucial when user and community interactions are subject to frequent changes.

Through extensive evaluation, we have shown that our approach outperforms baseline models in terms of recommendation quality, particularly for key industrial metrics at rank $K=5$. Additionally, it achieves improved scalability while preserving important topological properties of the original full graph.
These results highlight the practical applicability of our method in real-world marketing campaigns, where maintaining both high quality recommendations and scalability are critical. The LPA‐only pipeline is immediately deployable in low‐latency environments, while the GNN‐enhanced option suits settings where higher predictive performance justifies extra training cost. Future work will investigate adaptive coarsening techniques to further optimize performance in dynamic environments.

\section{Related Work}
\subsection{Graph Coarsening} Graph coarsening is widely used across industries primarily because it offers a practical way to distill massive graphs into manageable representations, enabling faster computations and clear understanding without losing the underlying connections.

Foundational to this is the multi-level paradigm posed by METIS \cite{karypis1997metis}, which uses techniques like heavy-edge matching to recursively simplify graphs for high-quality partitioning. The Louvain method \cite{blondel2008fast} performs coarsening by grouping densely connected nodes into communities. This approach has been further refined by algorithms like the Leiden method \cite{traag2019louvain}. Lean Algebraic Multigrid (LAMG) \cite{livne2012lean} leverages advanced numerical techniques for coarsening that maintains spectral fidelity, while \cite{loukas2018spectrally} introduced frameworks for spectrally approximating large graphs with smaller counterparts.

However, the most recent advancements are concentrated in learnable graph coarsening for Graph Neural Networks (GNNs). \cite{ying2018hierarchical} introduced a differentiable method to learn hierarchical cluster assignments for pooling nodes. Starting from this, many alternative approaches such as the node selection strategy \cite{gao2019graph}, Self-Attention Graph Pooling (SAGPool) \cite{lee2019self} has been implemented.

\subsection{Label Propagation Algorithms} LPAs are extensively utilized in industry because they offer a remarkably efficient and intuitive method for inferring missing information or uncovering latent structure within large-scale graphs. In original LPA, \cite{raghavan2007near}, nodes in the network keep copying the most common label from their neighbors. 

More recently, GNNs have become very popular. GNNs can be seen as a smart, learnable way to do label propagation. The general idea of message passing \cite{gilmer2017neural}, or attention-based GNN \cite{velivckovic2017graph}, also work by passing information between neighbors. These deep learning models have achieved top results on many graph tasks. However, in \cite{huang2020combining}, it is demonstrated that classical label propagation can work as well as, or even better than, complex GNNs. This is especially true when there isn't much labeled data, or when speed is fundamental. This key finding shows that simple propagation methods are still very powerful and efficient.

\subsection{GNNs in Industrial Recommender Systems}
GNNs have become fundamental in modern industrial recommender systems. Pinterest's Pinsage \cite{ying2018graph} was a seminal work demonstrating how GNNs, leveraging random walks and graph convolutions, could learn recommendations for billions of items. Alibaba further pushed the boundaries with algorithms \cite{wang2018billion, wang2020m2grl} designed to handle large-scale heterogeneous graphs.
Furthermore, foundational advancements in GNN architectures, such as Graph Attention Networks (GAT) \cite{velivckovic2017graph}, have provided powerful mechanism for modeling complex relational data, significantly influencing the subsequent design of GNN-based recommendations systems across the industry. Leveraging the power of such graph learning architectures, Linkedin has developed models like LiGNN \cite{borisyuk2024lignn}, which refines recommendation within its vast graph.

\section{GenAI Usage Disclosure}
In accordance with the ACM's guidelines on the use of generative AI tools, we confirm that generative AI technologies were used solely to assist with code debugging and grammar checking when preparing this paper. All research ideas, experiments, analyses and writing were conducted and critically reviewed by the authors. No part of the scientific content or creative reasoning was generated or substantially rewritten by generative AI tools.

\bibliography{bibliography}

\end{document}